\documentclass{article}
\usepackage{spconf,amsmath,graphicx}
\usepackage{booktabs}
\usepackage{hyperref}
\usepackage{xcolor}

\newcommand{\etal}{\textit{et al. }}

\title{Speech Emotion Recognition with Distilled Prosodic and Linguistic Affect Representations}

\name{Debaditya Shome, Ali Etemad}
\address{Queen's University, Canada}

\begin{document}
%
\maketitle
\begin{abstract}
We propose EmoDistill, a novel speech emotion recognition (SER) framework that leverages cross-modal knowledge distillation during training to learn strong linguistic and prosodic representations of emotion from speech. During inference, our method only uses a stream of speech signals to perform unimodal SER thus reducing computation overhead and avoiding run-time transcription and prosodic feature extraction errors. During training, our method distills information at both embedding and logit levels from a pair of pre-trained Prosodic and Linguistic teachers that are fine-tuned for SER. Experiments on the IEMOCAP benchmark demonstrate that our method outperforms other unimodal and multimodal techniques by a considerable margin, and achieves state-of-the-art performance of $77.49$\% unweighted accuracy and $78.91$\% weighted accuracy. Detailed ablation studies demonstrate the impact of each component of our method. 
\end{abstract}


\section{Introduction}
\label{sec:intro}
Speech Emotion Recognition (SER) is a challenging yet crucial task, with applications spanning a broad spectrum from human-computer interaction to mental health diagnostics. The inherent ambiguity in perceiving emotions and the variability across speakers and languages further amplifies the complexity of SER. 

Speech emotion information is present in and can be extracted from two different domains, linguistic and prosodic. The linguistic information includes the semantic aspects of emotion at the word level, while prosodic information includes the melodic aspects such as rhythm, tone, pitch, pauses, etc. Most existing solutions attempt to implicitly learn a combination of the two domains directly from raw speech signals. However, we identify four key problems in this category of approaches as follows: \\
\noindent (\textit{i}) Implicitly learning prosodic information from audio is often less than optimal because the discretization of audio signals during training of leading speech models like HuBERT \cite{hsu2021hubert} and Wav2Vec2 \cite{baevski2020wav2vec} can lead to the weakening of important prosodic features.\\
\noindent (\textit{ii}) Direct fine-tuning of existing speech models which were originally trained for Automatic Speech Recognition (ASR), on SER tasks, may not always yield strong performances \cite{pepino2021emotion}. \\
(\textit{iii}) Direct use of speech transcripts at \textit{run-time} can lead to low performances due to transcription errors \cite{li2022fusing}. \\
(\textit{iv}) Lastly, the use of both audio and linguistic information at run-time requires a multimodal system which can increase computational overhead.

\begin{figure*}[htbp!]
    \centering
    \includegraphics[width=0.76\textwidth]{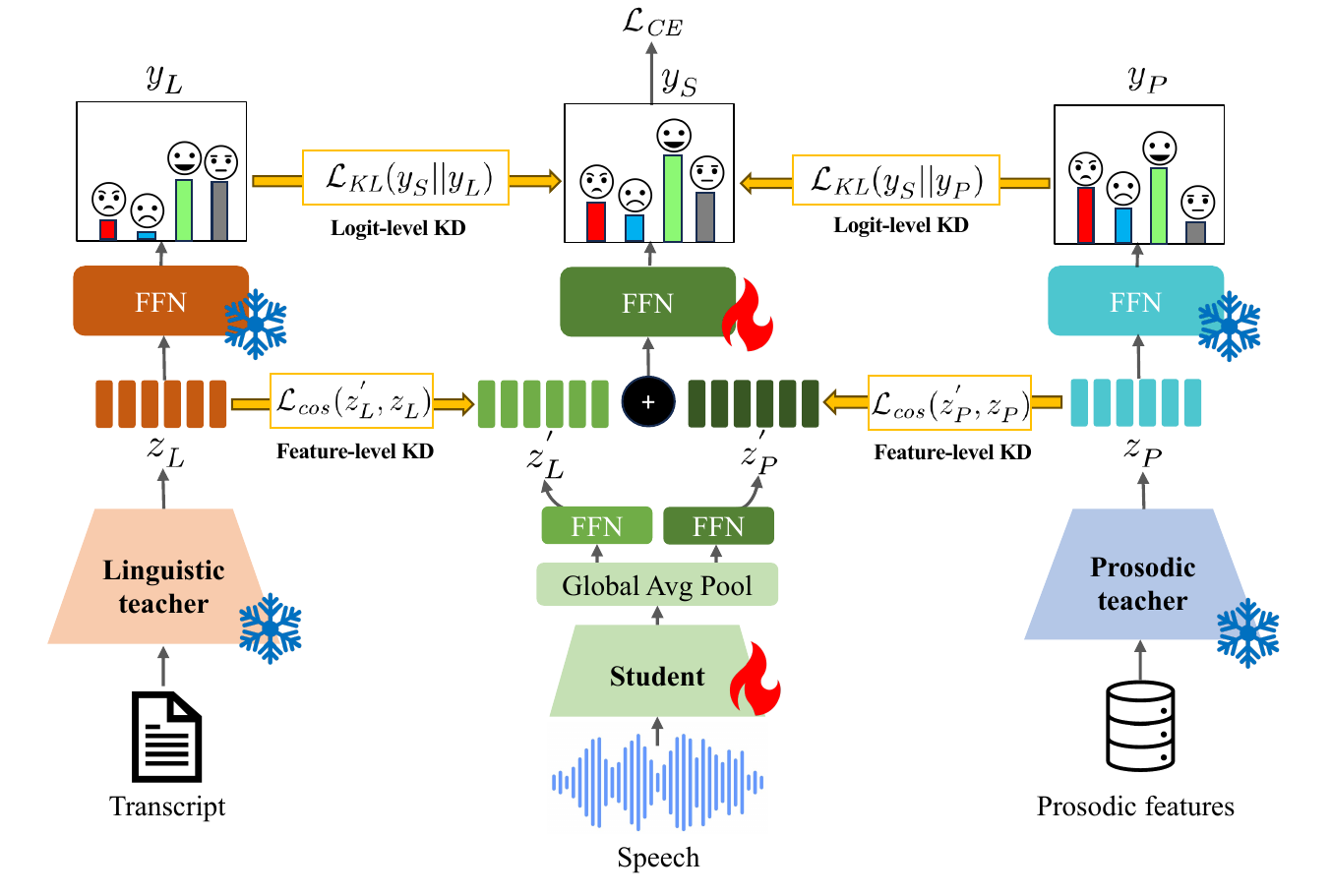}
    \caption{EmoDistill Framework. Our student network is trained using a distillation of logit-level and embedding-level knowledge from frozen linguistic and prosodic teacher networks, along with standard cross-entropy loss. During inference, we only use the student network in a unimodal setup, avoiding computational overhead as well as transcription and prosodic feature extraction errors.}
    \label{fig:framework}
\end{figure*}

To tackle the problems stated above, we propose EmoDistill, an SER method that learns both prosodic and linguistic information during training, but requires only input speech at run-time. Our method distills information from both logits and embeddings through a pre-trained prosodic teacher alongside a pre-trained linguistic teacher to learn unimodal representations for downstream SER. Experiments demonstrate that our method significantly outperforms prior solutions on the IEMOCAP \cite{busso2008iemocap} dataset to achieve state-of-the-art results. Additionally, ablation studies demonstrate the importance of each component of EmoDistill.

In summary, we make the following contributions:
(\textbf{1}) We introduce EmoDistill, a novel cross-modal Knowledge Distillation (KD) framework for learning unimodal representations from speech that explicitly capture both the linguistic and prosodic aspects of emotions. Unlike multimodal models combining audio and text modalities, EmoDistill doesn't require explicitly transcribed text during inference, thereby reducing the computational overhead and errors that arise from transcription and prosodic feature extraction.
(\textbf{2}) We empirically evaluate the importance of the ability to capture and distinguish linguistic and prosodic components of emotion in speech through detailed ablation studies.
(\textbf{3}) Our rigorous evaluation on the IEMOCAP benchmark in a subject-independent setup demonstrates that EmoDistill outperforms previous state-of-the-art methods and achieves $77.49$\% unweighted accuracy (UA) and $78.91$\% weighted accuracy (WA).

\section{Related Work}

The recent progress of deep learning has had a considerable impact on the field of SER. Mao \etal \cite{mao2014learning} utilized a Convolutional Neural Network (CNN) with Autoencoder-based pre-training for improved SER performance.  Luo \etal \cite{luo2018investigation} explored the combination of handcrafted features and Convolutional Recurrent Neural Network (CRNN) architecture for SER. Similarly, different variants of CNNs, RNNs, or CRNNs have been developed for SER, some of which have been equipped with Attention mechanisms. Recently, transformer-based speech models with self-supervised pre-training have shown promising performance in various downstream tasks including SER. Wang \etal \cite{wang2021fine} fine-tuned several variants of HuBERT and Wav2Vec2 for SER, speaker verification, and spoken language understanding tasks. Wagner \etal \cite{wagner2023dawn} analyzed the various factors like fairness, generalization, efficiency, and robustness of pre-trained speech models for continuous SER. They found that such pre-trained transformers show better robustness and fairness compared to CNNs. 

Multimodal methods that incorporate both speech and text in training and run-time have also been explored for SER.  Sun \etal \cite{sun2021multimodal} utilized CNN and CNN-LSTM networks for multimodal SER from speech and text data on the IEMOCAP corpus. Heusser \etal \cite{heusser2019bimodal} explored multimodal fusion from BiLSTM-based speech features with text features from a pre-trained XLNet language model. Triantafyllopoulos \etal \cite{triantafyllopoulos2023multistage} studied various combinations of speech features from Multi-stage CNNs and text features from BERT for SER and demonstrated improved performance. 
Deschamps \etal \cite{deschamps2023exploring} analyzed several multimodal fusion strategies using Wav2Vec2-based speech features and FlauBERT-based text features for SER on an emergency call-center recordings corpus. Ho \etal \cite{ho2020multimodal} proposed an SER method with a multi-level multi-head attention mechanism for the fusion of MFCC-based audio features and BERT-based text features.

As discussed in the previous section, fusion-based methods have multiple disadvantages like transcription errors, due to which cross-modal KD is being explored. KD was introduced by Hinton \etal \cite{hinton2015distilling} for model compression, where they utilized only logit-level information. Subsequently, KD was adapted to transfer cross-modal information in low-resource tasks such as SER. Hajavi \etal \cite{hajavi2023audio} used video as privileged information for distilling feature-level knowledge into a unimodal student on speech data, and demonstrated improved performance on speaker recognition and SER. Ren \etal \cite{ren2023fast} developed a self-distillation framework for SER aimed at model compression and demonstrated improvements over layer-wise KD.

\section{Method} 
\label{sec:method}

The objective of our framework is to train an unimodal speech student model using KD from pre-trained prosodic and linguistic teacher models. The overview of our method is presented in Figure \ref{fig:framework}. The details of each component are described as follows.

\noindent\textbf{Linguistic teacher.}
We consider a teacher model $f^{L}_{T}$ with strong language representations and refer to it as Linguistic teacher. We adopt the pre-trained \textit{BERT-base} \cite{devlin2018bert} model as the backbone for $f^{L}_{T}$, and perform supervised fine-tuning on the training set of our emotion classification corpus.

\noindent\textbf{Prosodic teacher.} We consider a teacher model $f^{P}_{T}$ that takes explicit prosodic features as input, and refer to it as Prosodic teacher. We use eGeMAPs Low-Level Descriptors (LLDs) \cite{eyben2015geneva} as prosodic features, which are commonly used in SER literature. We perform supervised fine-tuning of $f^{P}_{T}$ on the training set of our emotion classification corpus. We adopt a 2D ResNet-based \cite{he2016deep} backbone for $f^{P}_{T}$ which consists of $4$ residual blocks. 

\noindent\textbf{Student KD.} To facilitate knowledge transfer from our Linguistic and Prosodic teacher models, we follow a teacher-student KD setup and keep the weights of the teachers frozen. We consider a uni-modal speech model $f_{S}$ as the student, which consists of a pre-trained transformer encoder followed by $2$ GELU-activated feedforward projection layers for disjoint linguistic and prosodic embeddings. We keep these disjoint to allow optimal embedding-level KD from each teacher without interference. These two embeddings are concatenated and passed on to a feed-forward network (FFN) for final output predictions. First, we transfer the logit-level knowledge using traditional KD with temperature-scaled labels \cite{hinton2015distilling}. Specifically, we minimize the KL-Divergence $L_{KL}$ between the predicted logit distributions of teacher and student models, where the objective becomes:
\begin{equation}
    \mathcal{L}_\text{\textit{logits}} = \mathcal{L}_\text{\textit{KL}}(y^{}_S || y^{}_L) + \mathcal{L}_\text{\textit{KL}}(y^{}_S || y^{}_P).
\end{equation}
Here, $y^{}_S$ refers to the predictions of the student, while $y^{}_L$ and $y^{}_P$ represent the predictions of Linguistic and Prosodic teacher models, respectively. In all cases, the predicted logits $y$ are obtained using temperature parameter $\tau$ in the output softmax activation function. In practice, we use different values of $\tau$ for KD from $f^{L}_{T}$ and $f^{P}_{T}$. Let $z_c$ be the output logits for class $c$, among a total of $N$ classes. The temperature-scaled logits $y_c$ are obtained as:
\begin{equation}
    y_c = \frac{e^{z_c/\tau}}
    {\sum_{k=1}^{N}e^{z_c/\tau}}.
    \label{eq:logits}
\end{equation}
Next, we use embedding-level KD to transfer knowledge to the student model from the latent space of Linguistic and Prosodic teacher models. Let $z^{}_L$ and $z^{}_P$ denote the embeddings of Linguistic and Prosodic teachers, while $z^{'}_L$ and $z^{'}_P$ denote the embeddings of the student model from linguistic and prosodic projection layers respectively. We minimize the negative cosine similarity $L_{cos}$ among the teacher and student embeddings as follows:
\newcommand{\lnorm}[1]{\frac{#1}{\left\lVert{#1}\right\rVert _2}}
\begin{equation}
    \mathcal{L}_\text{\textit{embeddings}} = \mathcal{L}_\text{\textit{cos}}(z^{'}_L, z^{}_L) + \mathcal{L}_\text{\textit{cos}}(z^{'}_P, z^{}_P).
\end{equation}
Given two embeddings $a$ and $b$, $L_{cos}$ can be defined as:
\begin{equation}
    \mathcal{L}_\text{\textit{cos}}(a, b) = \lnorm{a}\cdot\lnorm{b},
\end{equation}
where ${\left\lVert{\cdot}\right\rVert _2}$ represents $\ell_2$-norm. Finally, the total training loss of EmoDistill becomes:
\begin{equation}
    \mathcal{L}_\text{\textit{EmoDistill}} = \alpha \mathcal{L}_\text{\textit{logits}} + \beta \mathcal{L}_\text{\textit{embeddings}} + \gamma \mathcal{L}_\text{\textit{CE}},
\end{equation}
where $\mathcal{L}_\text{\textit{CE}}$ refers to the standard cross-entropy loss, and $\alpha$, $\beta$, $\gamma$ are loss coefficients. 
 
\section{Experiments}
\label{sec:setup}
\subsection{Dataset}
We use the Interactive Emotional Dyadic Motion Capture (IEMOCAP) dataset for our experiments \cite{busso2008iemocap}. IEMOCAP is the most widely used benchmark for SER.
The dataset encompasses roughly $12$ hours of audio-visual content, with an average duration of $4.5$ seconds for each vocal segment. We only use the audio and text transcriptions in this work. Following prior works, we use 4 categories of emotions: `neutral', `angry', `sad', and `happy' (merged with `excited' class). 

\subsection{Implementation details}
We train all models on $4 \times$NVIDIA A100 GPUs, using a batch size of 128, except for EmoDistill w/HuBERT-large for which we use batch size of 64 due to computational limitations. We use AdamW 
optimizer with $\mathrm{CosineAnnealingWarmup}$ learning rate (LR) scheduler starting with a base LR of $1\times10^{-4}$. For logit-level KD from the Prosodic teacher, a temperature $\tau_{P} = 0.5$ is chosen, while for the Linguistic teacher temperature $\tau_{L} = 5$ (see ablation experiments in Section \ref{sec:results}). $\alpha = 1$, $\beta = 10$, $\gamma = 2$ are used as loss coefficients. The pre-trained weights of HuBERT-base and HuBERT-large were obtained from TorchAudio. 
For BERT-base, we use the `bert-base-uncased' checkpoint from HuggingFace. 
For extracting eGeMAPs LLDs, we use the opensmile-python toolkit \cite{eyben2010opensmile}.

\subsection{Results and Discussion}
\label{sec:results}
\textbf{Performance.} Following prior works, we evaluate EmoDistill on the IEMOCAP benchmark using $10$-fold cross-validation in the leave-one-speaker-out scheme. The results are shown in Table \ref{table:leave-one-speaker}. It can be clearly seen that EmoDistill significantly outperforms prior works in terms of both WA and UA metrics, with improvements of up to $7.26\%$ in UA and $4.99\%$ in WA over the best previous method \cite{ye2023temporal}. Furthermore, we observe that while our method is technically not multi-modal as it only uses a single modality during inference, it still outperforms prior works that have dedicated components for different text and audio modalities in the literature \cite{heusser2019bimodal, sun2021multimodal, ho2020multimodal, triantafyllopoulos2023multistage}.

\textbf{Ablation studies.} To understand the impact of each component of EmoDistill, we conduct a systematic ablation study and present the results in Table \ref{table:ablation}. First, we individually remove the $\mathcal{L}_\textit{\text{logits}}$ as well as $\mathcal{L}_\textit{\text{embedding}}$ and observe between 1\% to 2\% drop in performance in each case. Next, we ablate the model by individually removing the entire Prosodic and Linguistic teachers ($f^{P}_{T}$ and $f^{L}_{T}$). In this experiment, we observe that while the removal of either component degrades performance, the ablation of the Linguistic teacher has a more significant negative impact. We then ablate both teachers together ($f^{P}_{T}$ and $f^{L}_{T}$), essentially only using the HuBERT-base backbone with fine-tuning for SER, and observe a considerable drop in performance. Finally we remove the student network along with either of the teachers, essentially only using the remaining teacher for inference. We observe here that while both tests result in a considerable drop in performance, the removal of $f_S$ and $f^{L}_{T}$ together has the highest negative impact, indicating  that linguistic information is crucial, and prosodic information can serve as complementary knowledge to improve SER but can't replace linguistic information.

\begin{figure}[t]
    \centering
    \setlength\tabcolsep{0pt}
    \begin{tabular}{cc}
    \includegraphics[width=0.482\linewidth]{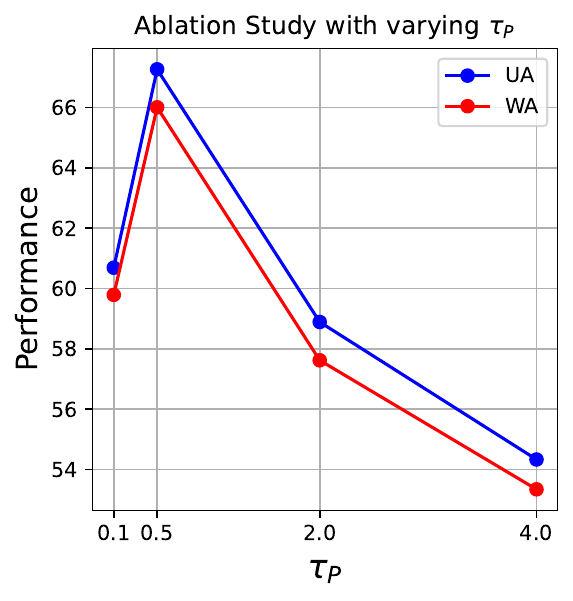}
         &  
    \includegraphics[width=0.5\linewidth]{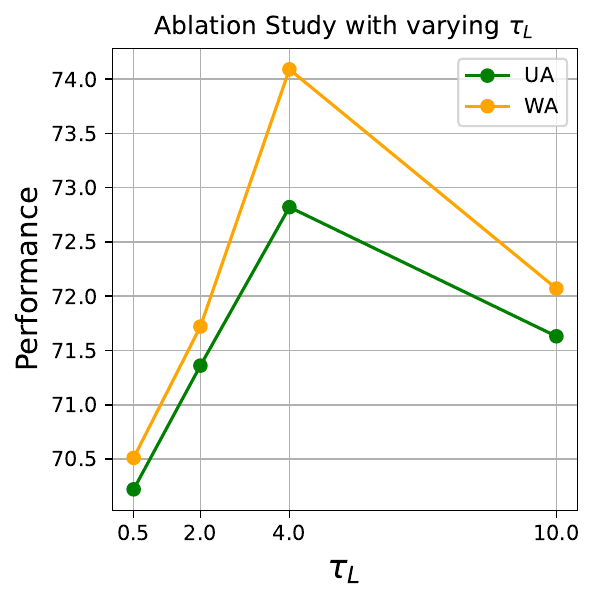}
    \\
    \end{tabular}
    \caption{
    \textbf{Left:} We remove $f^{L}_{T}$ and vary $\tau_{P}$.  
    \textbf{Right:}  We remove $f^{P}_{T}$ and vary $\tau_{L}$. 
    }
    \label{fig:ablation}
\end{figure}

Next, we aim to analyze the impact of the temperature parameter $\tau$ on the performance. To this end, remove $f^{L}_{T}$ and set the prosodic temperature parameter $\tau_{P}$ to $0.1$, $0.5$, $2$, and $4$. Similarly, we remove $f^{P}_{T}$ and set the linguistic temperature parameter $\tau_{L}$ to $0.5$, $2$, $4$, and $10$. As shown in Figure \ref{fig:ablation} (Left), $\tau_{P} = 0.5$ (hard-logits) works best and increasing $\tau_{P}$ shows strong decline in performance. In the second case, as shown in Figure \ref{fig:ablation} (Right), we observe that $\tau_{L} = 4$ (soft-logits) works best and decreasing $\tau_{L}$ leads to a strong decline in performance. Although standard logit-level KD methods use soft logits ($\tau > 1$), we observe that soft logits don't work well for the Prosodic teacher $f^{P}_{T}$. Our intuition is that since $f^{P}_{T}$ is a weak teacher (see Table \ref{fig:ablation}), smaller temperature values result in hard logits as per Eq. \ref{eq:logits}, and therefore improve performance by providing stronger supervision signals through distillation.
Finally, we observe that for both teachers, too high or low temperatures lead to a drop in performance.

\begin{table}[t]
\centering
\setlength{\tabcolsep}{2.5pt}
\small
\caption{SER results on IEMOCAP. \textbf{Bold} denotes the best results while \underline{underline} denotes the second-best.}
\begin{tabular}{lcccc}
\toprule
\textbf{Method} & \textbf{Inf. Backbone} & \textbf{Modality} & \textbf{WA} & \textbf{UA}\\\midrule \midrule
\textcolor{gray}{\cite{sun2021multimodal}} & \textcolor{gray}{CNN+LSTM} & \textcolor{gray}{Multimodal} & \textcolor{gray}{61.2} & \textcolor{gray}{56.01} \\
\textcolor{gray}{\cite{heusser2019bimodal}}& \textcolor{gray}{BiLSTM+XLNet} & \textcolor{gray}{Multimodal} & \textcolor{gray}{71.40}&\textcolor{gray}{68.60}\\
\textcolor{gray}{\cite{triantafyllopoulos2023multistage}} & \textcolor{gray}{MFCNN+BERT} & \textcolor{gray}{Multimodal} & \textcolor{gray}{-} & \textcolor{gray}{72.60}\\
\textcolor{gray}{\cite{ho2020multimodal}} & \textcolor{gray}{RNN+BERT} & \textcolor{gray}{Multimodal} & \textcolor{gray}{73.23} & \textcolor{gray}{74.33}\\
\midrule
\cite{aftab2022light}& FCNN & Unimodal & 70.23 & 70.76\\
\cite{liu2020speech}& TFCNN+DenseCap+ELM & Unimodal &70.34&70.78\\
\cite{cao2021hierarchical}& LSTM+Attention& Unimodal & 70.50 & 72.50\\
\cite{lu2020speech}& RNN-T & Unimodal &71.72&72.56\\
\cite{wu2021speech}& CNN-GRU+SeqCap & Unimodal &72.73&59.71\\
\cite{zou2022speech} & Wav2Vec2+CNN+LSTM & Unimodal & 71.64 & 72.70\\
\cite{ye2023temporal}& TIM-Net & Unimodal &72.50&71.65\\
\midrule
Ours & HuBERT-base & Unimodal & \underline{75.16} & \underline{76.12}\\
Ours & HuBERT-large & Unimodal & \textbf{77.49} & \textbf{78.91}\\
\bottomrule
\end{tabular}
\label{table:leave-one-speaker}
\end{table}

\begin{table}[t]
\centering
\small
\caption{Ablation study demonstrating the impact of key components of EmoDistill.}
\begin{tabular}{lcc}
\toprule
\textbf{Variants} & \textbf{WA} & \textbf{UA}\\\midrule \midrule
Ours & \textbf{75.16} & \textbf{76.12}\\
w/o $\mathcal{L}_\textit{\text{logits}}$ & 73.94 $(\downarrow 1.22)$ & 74.02 $(\downarrow 2.10)$\\
w/o $\mathcal{L}_\textit{\text{embedding}}$ & 73.88 $(\downarrow 1.28)$ & 74.01 $(\downarrow 2.11)$\\
w/o $f^{P}_{T}$ & 74.09 $(\downarrow 1.07)$ & 72.82 $(\downarrow 3.30)$\\
w/o $f^{L}_{T}$ & 66.01 $(\downarrow 9.15)$ & 67.27 $(\downarrow 8.85)$ \\
w/o $f^{P}_{T}$ and $f^{L}_{T}$ & 69.92 $(\downarrow 5.24)$ & 70.17 $(\downarrow 5.95)$\\
w/o $f_S$ and $f^{L}_{T}$ & 49.42 $(\downarrow 25.74)$ & 50.08  $(\downarrow 26.04)$\\
w/o $f_S$ and $f^{P}_{T}$ & 71.09 $(\downarrow 4.07)$ & 71.83 $(\downarrow 4.29)$\\
\bottomrule
\end{tabular}
\label{table:ablation}
\end{table}

\section{Conclusion}
We present EmoDistill, a novel cross-modal knowledge distillation framework for learning emotion representations from speech. EmoDistill explicitly captures linguistic and prosodic aspects of emotions in a unimodal inference setup, reducing computational overhead and limitations like transcription and prosodic feature extraction errors. For training our framework, EmoDistill extracts information from both the embedding and logit levels through a pair of pre-trained Prosodic and Linguistic teacher models that have been fine-tuned for SER. Experiments on the commonly used SER benchmark IEMOCAP demonstrates that our method considerably outperforms other state-of-the-art methods by achieving $77.49\%$ ($7.26\%$ improvement) and $78.91\%$ ($4.99\%$ improvement) weighted and unweighted accuracies. We demonstrate the importance each component of our method through detailed ablation experiments.
Practical applications of EmoDistill may include scenarios with low compute resources, and where emotions are expressed not only through language and semantics but also through prosody.

\section*{Acknowledgement}
This work was supported by Mitacs, Vector Institute, and Ingenuity Labs Research Institute.

\bibliographystyle{IEEEbib}
\small
\bibliography{refs}

\end{document}